\documentclass[conference]{IEEEtran}
\IEEEoverridecommandlockouts
\usepackage{cite}
\usepackage{amsmath,amssymb,amsfonts}
\usepackage{algorithmic}
\usepackage{graphicx}
\usepackage{textcomp}
\usepackage{xcolor}
\usepackage{multirow}
\usepackage{booktabs}
\def\BibTeX{{\rm B\kern-.05em{\sc i\kern-.025em b}\kern-.08em
    T\kern-.1667em\lower.7ex\hbox{E}\kern-.125emX}}
\begin{document}

\title{An Effective Contextual Language Modeling Framework for Speech Summarization with Augmented Features
}

\author{\IEEEauthorblockN{Shi-Yan Weng}
\IEEEauthorblockA{\textit{National Taiwan Normal University} \\
Taipei, Taiwan \\
40547041S@ntnu.edu.tw}
\and
\IEEEauthorblockN{Tien-Hong Lo}
\IEEEauthorblockA{\textit{National Taiwan Normal University} \\
Taipei, Taiwan \\
teinhonglo@ntnu.edu.tw}
\and
\IEEEauthorblockN{Berlin Chen}
\IEEEauthorblockA{\textit{National Taiwan Normal University} \\
Taipei, Taiwan \\
berlin@ntnu.edu.tw}
}

\maketitle

\begin{abstract}
Tremendous amounts of multimedia associated with speech information are driving an urgent need to develop efficient and effective automatic summarization methods. To this end, we have seen rapid progress in applying supervised deep neural network-based methods to extractive speech summarization. More recently, the Bidirectional Encoder Representations from Transformers (BERT) model was proposed and has achieved record-breaking success on many natural language processing (NLP) tasks such as question answering and language understanding. In view of this, we in this paper contextualize and enhance the state-of-the-art BERT-based model for speech summarization, while its contributions are at least three-fold. First, we explore the incorporation of confidence scores into sentence representations to see if such an attempt could help alleviate the negative effects caused by imperfect automatic speech recognition (ASR). Secondly, we also augment the sentence embeddings obtained from BERT with extra structural and linguistic features, such as sentence position and inverse document frequency (IDF) statistics. Finally, we validate the effectiveness of our proposed method on a benchmark dataset, in comparison to several classic and celebrated speech summarization methods.
\end{abstract}


\begin{figure*}
\centering
\includegraphics[width=10cm]{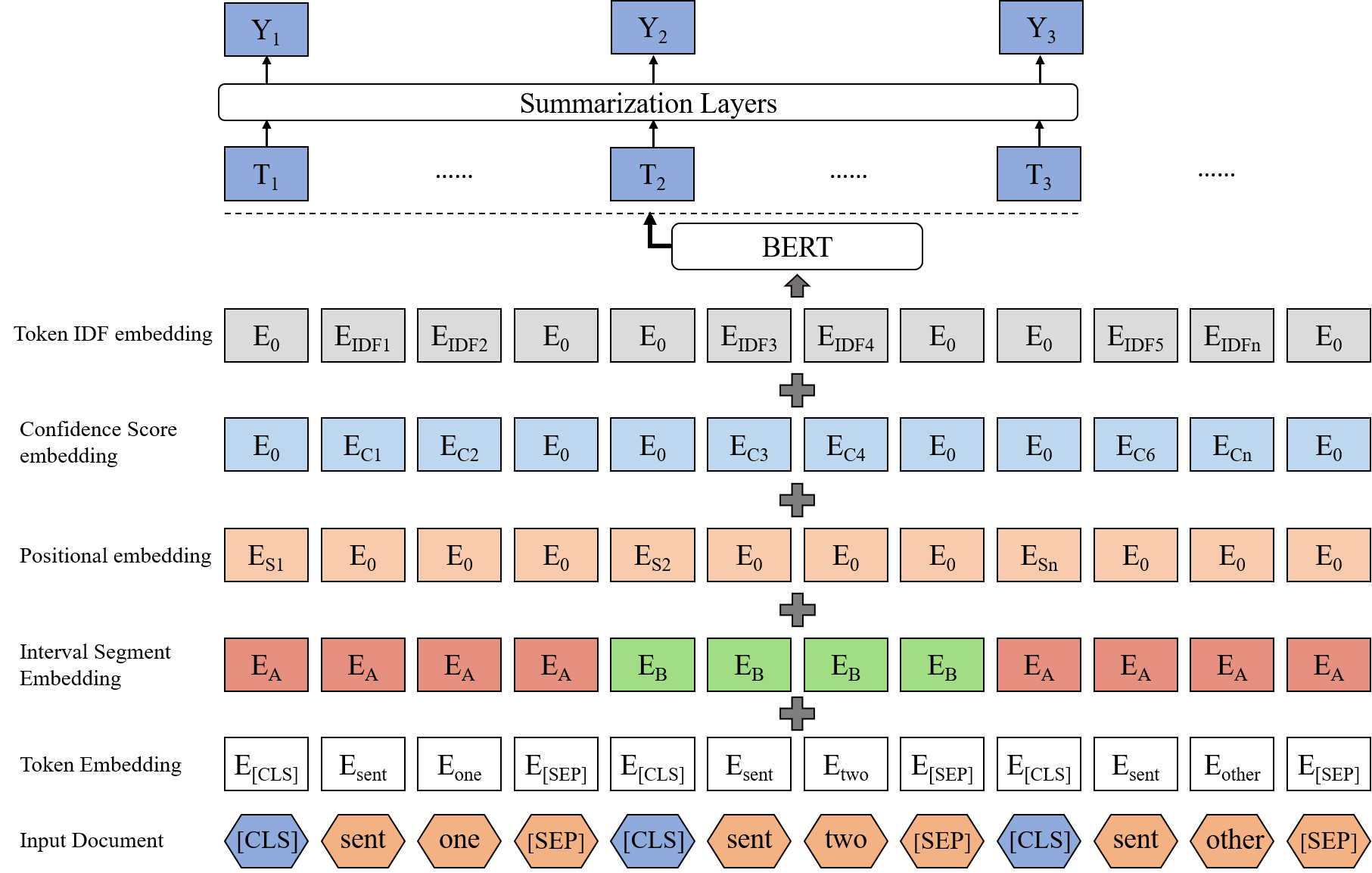}
\caption{The architecture of the proposed method}
\label{fig:picture001}
\end{figure*}

\section{Introduction}
Along with the popularity of the Internet, the exponential growth in the volumes of multimedia associated with speech content available on the Web, such as radio broadcasts, television programs, online MOOCs, lecture recordings and among others, has necessitated the development of effective and efficient summarization techniques. For example, extractive speech summarization, which seeks to select the most important and representative sentences from a source spoken document so as to create a short summary, can help users quickly navigate and digest multimedia content by listening to the corresponding speech segments of the summary. However, extractive speech summarization may be inevitably vulnerable the detrimental effect of recognition errors when using automatic speech recognition (ASR) techniques to transcribe spoken documents into text form. Simultaneously, in recent years we have seen a significant body of work devoted to the exploration of unsupervised and supervised deep neural network-based methods for extractive text summarization. The main concept lies behind the former methods is to learn continuously distributed (as opposed to one-hot) vector representations of words using neural networks in an unsupervised manner. The learned representations are anticipated to encode word-level synonymy relations and proximity information, which in turn can be used to infer similarity or relevance among words, sentences and documents. A common thread of leveraging such word embedding methods in extractive text summarization is to represent the document to be summarized and each of its constituent sentences by averaging the corresponding word embeddings over all words within them, respectively. After that, the cosine similarity measure, as a straightforward choice, can be readily applied to determine the salience (or relevance) of each sentence with regard to the document. On the other hand, a general thought of the methods in the latter category is to conceptualize extractive summarization as a sequence label problem [1-9], where each sentence of a document to summarized is quantified with a score (or tagged with a label) that facilitate determine whether the sentence should be included in the summary or not. Most of the cutting-edge instantiations typically follow a two-step strategy. First, a recurrent neural network (RNN)-based encoder is employed to obtain a holistic representation of the document by taking the representations of its constituent sentences as the inputs to RNN successively. Second, an RNN-based decoder that takes the document representation as the initial input is then used to quantify (or label) each sentence in tandem, meanwhile taking the previously processed sentences into account. Yet, there still are a wide array of classic unsupervised summarization methods, such as LEAD (the leading-sentences method) and ILP (Integer Linear Programming), which are intuitively simple and can achieve competitive performance on several popular benchmark summarization tasks in relation to the deep neural network-based methods. The interested reader is referred to [10] for a comprehensive overview of the classic unsupervised summarization methods. More recently, the so-called Bidirectional Encoder Representations from Transformers (BERT) [11] model, a novel neural network-based contextual language model, has shown very impressive results on many natural language processing (NLP) tasks like question answering, language understanding.

Based on the above observations, this paper presents a continuation and extension of the BERT-based method for supervised extractive speech summarization, while making the following contributions. First, we propose a novel BERT-based extractive summarization framework, which manages to robustly perform summarization on spoken documents equipped with erroneous ASR transcripts. Second, we explore the use of several auxiliary structural and linguistic features to enrich the embeddings of the sentences of a spoken document in order to further promote the summarization performance. Finally, we carry out extensive sets of experiments to demonstrate the effectiveness of the summarization methods stemming from our modeling framework, by comparing them with several strong baselines on a benchmark dataset.

The remainder of this paper is organized as follows. We first describe the fundamentals of the BERT-based method in Section 2. In Section 3, we shed light on our proposed novel BERT-based modeling framework for speech summarization. After that, the experimental setup and results are presented in Section 4. Section 5 concludes this paper and presents possible avenues for future research.

\section{Speech Summarization With Bert}

\subsection{Bidirectional Encoder Representations from Transformers (BERT)}
BERT is an innovative neural language model which makes effective use of bi-directional self-attention (also called the Transformer) to capture both short- and long-span contextual interaction between the tokens in its input sequences, usually in the form of words or word pieces. In contrast to the traditional word representation such as word2vec or GLOVE, the advantage of BERT is that it can produce context-aware representation for the same word at different locations by considering bi-directional dependency relations of words across sentences. The training of BERT consists of two stages: pre-training and fine-tuning. At the pre-training stage, its model parameters can be estimated on huge volumes of unlabeled training data over different tasks such as the masked language model task and the next (relevant) sentence prediction task [11]. At the fine-tuning stage, the pre-trained BERT model, stacked with an additional single- or multi-layer perceptron (MLP), can be fine-tuned to work well on many NLP-related tasks when only a very limited amount of supervised task-specific training data are made available. A bit more terminology: we will explain below a possible usage of BERT for spoken document summarization. First, for the masked language model task conducted at the pre-training stage, given a token (e.g., word or character) sequence $\mathbf{x}=[x_1,...,x_n]$, BERT constructs $\hat{\mathbf{x}}$ by randomly replacing a proper portion of tokens in $x$ with a special symbol \verb|[mask]| for each of them, and designates the masked tokens collectively be $\Bar{\mathbf{x}}$. Let $H_\theta$ denotes a Transformer which maps a length-$T$ token sequences $\mathbf{x}$ into a sequence of hidden vectors $H_\theta(\mathbf{x})=[H_\theta(x)_1, H_\theta(x)_2,..., H_\theta(x)_n]$, then the pre-training objective function of BERT [30] can be expressed by:
\begin{equation}
\begin{aligned}
    \max_\theta \mathrm{log}p_\theta(\Bar{\mathbf{x}}|\Tilde{\mathbf{x}})&\approx \sum_{t=1}^T m_t \mathrm{log}p_\theta (x_t|\hat{\mathbf{x}})\\
    &=\sum_{t=1}^T m_t\mathrm{log}\frac{\mathrm{exp}(H_\theta(\hat{x})^\intercal_t e(x_t))}{\sum_{x^\prime} \mathrm{exp}(H_\theta(\hat{x})^\intercal_t) e(x^\prime)}
\end{aligned}
\end{equation}
where $m_t$ is an indicator of whether the token at position $t$ is masked or not.

\begin{table*}[]
\renewcommand\arraystretch{1.25}
\centering
\textbf{Table 2.}~~ Summarization results of the classic models and the strong baselines.\\
\setlength{\tabcolsep}{7mm}
\begin{tabular}{ccccccc}
\toprule[1.25pt]
\multirow{2}{*}{Methods} & \multicolumn{3}{c}{\textbf{Text Documents (TD)}}          & \multicolumn{3}{c}{\textbf{Spoken Documents (SD)}}        \\ \cline{2-7} 
                         & ROUGE-1        & ROUGE-2        & ROUGE-L        & ROUGE-1        & ROUGE-2        & ROUGE-L        \\ \midrule[1 pt]
VSM                      & 0.347          & 0.228          & 0.290          & 0.342          & 0.189          & 0.287          \\ \hline
LSA                      & 0.362          & 0.233          & 0.316          & 0.345          & 0.201          & 0.301          \\ \hline
SG                       & 0.410          & 0.300          & 0.364          & 0.378          & 0.239          & 0.333          \\ \hline
CBOW                     & 0.415          & 0.308          & 0.366          & \textbf{0.393} & \textbf{0.250} & \textbf{0.349} \\ \hline
DNN                      & 0.488          & 0.382          & 0.444          & 0.371          & 0.233          & 0.332          \\ \hline
CNN                      & \textbf{0.501} & \textbf{0.407} & \textbf{0.460} & 0.370          & 0.208          & 0.312          \\ \hline
Refresh                  & 0.453          & 0.372          & 0.446          & 0.329          & 0.197          & 0.319          \\ 
\bottomrule[1.25pt]
\end{tabular}
\end{table*}

\subsection{BERT-based summarization model}
As above mentioned, BERT has achieved considerable performance gains in almost all types of natural language processing applications, including question answering (QA) [12, 13], information retrieval (IR) [14, 15], dialog modeling, and others. Very recently, BERT has made inroads into extractive text summarization for use in identify salient summary sentences [16-20]. On the task of summary sentence selection, BERTSUM [16] treats a given document D to be summarized as a collection of sentences:
\begin{equation}
    D=[sent_1, sent_2,...,sent_m]
\end{equation}

At the input stage, BERTSUM adds \verb|[CLS]| at the beginning of each sentence and \verb|[SEP]| at the end of the sentence. In addition, as with the seminal paper of BERT, each sentence will finally be encoded with an interval segment vector to indicate change of sentence, while the vector for each word (token) is the concatenation of the position embedding, word embedding, interval segment embedding. The intermediate goal of BERTSUM is to use the output of the $i^th$ \verb|[CLS]| tag in the last layer of the BERT as the representation of each sentence $sent_i$ (denoted by $T_i$ for short). When training an extractive digestor, BERTSUM will output the representation of each sentence $sent_i$, encoded by BERT as $T_i$, to a classifier that determines whether $sent_i$ should be included into the summary or be excluded from it. If $sent_i$ is a summary sentence, the value $Y_i$, estimated by classifier, of which should get closer to 1 and 0 otherwise. Several classifiers may serve this purpose, including the simple classifier, Transformer, recurrent neural network (RNN) [21] and the like. As a side note, the objective of binary cross entropy can be in turn used to fine-tune the entire model components of BERTSUM.

\section{An Enhanced Bert-Based Method For Speech Summarization}
In this section, we will describe several novel extensions to make the BERT-based ranking model (summarizer) more suitable for dealing with spoken documents.
\subsection{Auxiliary Embedding}
To incorporate more spoken document-related knowledge into the BERT-based ranking model, we propose in this paper a straightforward yet effective approach that appends extra information cues to the embeddings of sentences involved in a spoken document be summarized, as shown in Figure 1.

\begin{table}[]
\renewcommand\arraystretch{1.25}
\centering
\textbf{Table 1.}~~ The statistical information of MATBN used in\\ the summarization experiments.\\
\begin{tabular}{ccc}
\toprule[1.25 pt]
                             & Training Set & Evaluation Set \\ \hline
Number of Doc.               & 185          & 20             \\ \hline
Avg. Num. of Sent. per Doc.  & 20           & 23.3           \\ \hline
Avg. Num. of words per Sent. & 17.5         & 16.9           \\ \hline
Avg. Num. of words per Doc   & 326.0        & 290.3          \\ \hline
Word Error Rate (WER\%)      & \multicolumn{2}{c}{23.7}      \\ \hline
Char. Error Rate (CER\%)     & \multicolumn{2}{c}{20.8}      \\ 
\bottomrule[1.25 pt]
\end{tabular}
\end{table}

\begin{table*}[]
\renewcommand\arraystretch{1.25}
\centering
\textbf{Table 3.}~~ Summarization results of our proposed methods with different model configurations.\\
\setlength{\tabcolsep}{4mm}
\begin{tabular}{ccccccc}
\toprule[1.25pt]
\multirow{2}{*}{\textbf{Method (BERT + summarization layer)}} & \multicolumn{3}{c}{\textbf{Text Documents (TD)}} & \multicolumn{3}{c}{\textbf{Spoken Documents (SD)}} \\ \cline{2-7} 
                                                              & ROUGE-1        & ROUGE-2        & ROUGE-L        & ROUGE-1         & ROUGE-2         & ROUGE-L        \\ \midrule[1 pt]
Simpler   classification (SC)                                 & 0.483          & 0.388          & 0.454          & 0.382           & 0.251           & 0.349          \\ \hline
Inter-sentence   Transformer (IT)                             & 0.489          & 0.381          & 0.459          & 0.383           & 0.254           & 0.354          \\ \hline
RNN                                                           & 0.485          & 0.385          & 0.458          & 0.383           & 0.249           & 0.350          \\ \hline
SC w/   Positional Embedding                                  & 0.498          & 0.410          & 0.469          & 0.396           & 0.259           & 0.355          \\ \hline
IT w/   Positional Embedding                                  & 0.501          & 0.413          & 0.475          & 0.402           & 0.263           & 0.354          \\ \hline
RNN w/   Positional Embedding                                 & 0.502          & 0.411          & 0.472          & 0.399           & 0.265           & 0.356          \\ \hline
SC w/   Confidence                                            & -              & -              & -              & 0.412           & 0.336           & 0.407          \\ \hline
IT w/   Confidence                                            & -              & -              & -              & 0.415           & 0.337           & 0.408          \\ \hline
RNN w/   Confidence                                           & -              & -              & -              & 0.414           & 0.336           & 0.407          \\ \hline
SC w/ IDF                                                     & 0.511          & 0.435          & 0.504          & 0.409           & 0.335           & 0.402          \\ \hline
IT w/ IDF                                                     & \textbf{0.513} & \textbf{0.437} & \textbf{0.507} & 0.408           & 0.336           & 0.402          \\ \hline
RNN w/ IDF                                                    & 0.510          & 0.435          & 0.502          & 0.409           & 0.333           & 0.403          \\ \hline
SC w/   Confidence + IDF                                      & -              & -              & -              & 0.428           & 0.342           & 0.411          \\ \hline
IT w/   Confidence + IDF                                      & -              & -              & -              & \textbf{0.431}  & 0.342           & \textbf{0.412} \\ \hline
RNN w/   Confidence + IDF                                     & -              & -              & -              & 0.430           & \textbf{0.343}  & 0.409          \\ 
\bottomrule[1.25pt]
\end{tabular}
\end{table*}

\begin{itemize}
\item \textbf{Inverse Document Frequency:} Inverse document frequency (IDF) [22] is a commonly-used lexical statistical feature with solid theoretical foundations in some classic IR models. While the unigram (term frequency) is computed on a per sentence basis (to emphasize tokens with a high frequency), IDF is computed over all the document collection (to penalize non-discriminative tokens). Motivated by the above observation, we seek to append the IDF score for every single input token to their original embeddings, which can be viewed as a kind of augmented embedding. As shown in Figure 1, the IDF embedding for each token n is denoted by $E_{IDF}$.
\item \textbf{Confidence Scores:} In the context of speech summarization, using imperfect ASR transcripts often leads to degraded performance. To tackle this issue, we explore to append the confidence score [23] of each token in an input sentence of the spoken document be summarized to its original embedding, which was computed a priori by an ASR system. The confidence score, in the form of posterior probability estimated by automatic speech recognition system, can be used to measure the potential correctness of a word. The quality of the confidence measure discussed in this paper is evaluated by collectively considering the normalized cross entropy (NCE) [31], the equal error rate (EER) and the detection error trade-off (DET) curve [32]. The confidence score used as an augmented embedding for each token n is denoted by $E_{Cn}$ in Figure 1.
\item \textbf{Positional embedding:} Many studies have pointed out that people tend to place be more vocal in the first half of the article when writing articles, especially in news articles, so there are many summarization models that will take the position information of the sentence in the article as a feature. In [24], the authors proposed several sentence-level positional representation setups on different tokens like \verb|[SEP]| or \verb|[CLS]| to enrich the embeddings generated by BERT for text summarization. We will select the top-performing one resulting from the above-mentioned embedding method to work in conjunction with the other embeddings as shown in Figure 1.
\end{itemize}

\section{Experiments}
\subsection{Experimental Dataset}
A series of empirical experiments are conducted on a Mandarin benchmark broadcast new (MATBN) corpus [25]. The MATBN dataset is publicly available and has been widely used to evaluate several natural language processing (NLP)-related tasks, including speech recognition [26], information retrieval [27] and summarization [2, 7-9]. A subset of 205 broadcast news documents compiled between November 2001 and August 2002 was reserved for the summarization experiments. Furthermore, since broadcast news stories often follow a relatively regular structure as compared to other speech materials like conversations, the positional information would play an important role in extractive summarization of broadcast news stories. We hence chose 20 documents, for which the generation of reference summaries is less correlated with the positional information (or the position of sentences) as the held-out test set to evaluate the general performance of the proposed summarization framework, while the other subset of 185 documents the training set alongside their respective human-annotated summaries for estimating the parameters of the various supervised summarization methods compared in the paper. Table 1 highlights some basic statistics about the spoken documents of the training and evaluation sets. Our recognizer is built with Kaldi toolkit [33] and a commonly-used [34] recipe.

For the assessment of summarization performance, we adopt the widely-used ROUGE [35] measure. All the experimental results reported hereafter are obtained by calculating the F-scores. The summarization ratio, defined as the ratio of the number of words in the automatic (or manual) summary to that in the reference transcript of a spoken document, was set to 10\% in this research.

\subsection{Experimental Results}
At the outset, we report on the summarization results of two classic unsupervised methods, viz. vector space model (VSM) and latent semantic analysis (LSA) [2, 28], as well as several well-practiced neural network-based methods, including three supervised ones, viz. deep neural network (DNN) [7], convolutional neural network (CNN) [7] and Refresh [4], and two unsupervised ones, viz. skip-gram (SG) [8, 29] and continuous bag-of-word (CBOW) [8, 29]. These models were re-implemented with the best configurations respectively found in the above studies. The corresponding summarization results of those methods are depicted in Table 2, where Text Documents (TD) denote the results obtained based on the reference transcripts of spoken documents and Spoken Documents (SD) denote the results with the erroneous speech recognition transcripts. Several observations can be made from Table 2. First, when in an unsupervised manner, the neural network-based methods (viz. SG and CBOW) always outperform the traditional vector-based methods (viz. VSM and LSA) for both the TD and SD cases. Second, the supervised summarizers, including DNN, CNN and Refresh, perform better than SG and CBOW in the TD case and most of them in the SD case.

In the second set of experiments, we evaluate the performance levels of various summarization models stemming from our proposed modeling framework (cf. Section 3). These major distinctions between these models are: 1) different kinds of the summarization layer being used, and 2) the use of an auxiliary embedding (positional embedding, IDF embedding, or confidence-score embedding) or not. The corresponding results are shown in Table 3. Note here that, except for the models listed in the first three rows, all the rest models shown in Table 3 are, by default, with the positional embedding. As can be seen from Table 3, most of our models either outperform or perform at comparable levels to the existing state-of-the-art models shown in Table 2, across different summarization scenarios and metrics. The summarization results illustrated in the first three row represent that without any auxiliary embedding, Inter-sentence Transformer (IT) is the best performing one (for both the TD and SD cases) among the three kinds of classifier selection for BERT, also confirming the modeling power of the multi-head attention mechanism of Transformers for capture both intra- and inter-sentence correlation patterns of words in the document to be summarized. In addition, the results shown in the Row 7 to 9 of Table 3 also reveal that the use of confidence score can improve speech summarization performance when with the erroneous recognition transcripts, which confirms the utility of such an attempt for alleviating the undesirable impact of imperfect recognition transcripts. Furthermore, it is striking that, for the TD case, IDF embedding can significantly improve the summarization performance of BERT, which is attributed to the ability of IDF embedding to incorporate global knowledge of term (token) usage into the BERT-based summarization model. Lastly, we have also evaluated our models on the CNN/DailyMail (a large-scale benchmark) text corpus, achieving quite good summarization results by integrating IDF and Positional features early mentioned. Here we, however, omit the detailed results because of space limitations.
\section{Conclusions and Futurework}
In this paper, we have presented an effective BERT-based neural summarization framework for spoken document summarization. Both IDF embedding and confidence-score embedding, as well as their fusion, have been investigated to work in conjunction with our summarization framework. The augmentation of these extra features into the proposed modeling framework indeed can further boost the summarization performance. In the future, we plan to develop more sophisticated neural networks, feature augmentation techniques and attention mechanisms for use in various tasks of speech summarization.


\end{document}